\documentclass{sn-jnl}
\usepackage{graphicx}%
\usepackage{multirow}%
\usepackage{amsmath,amssymb,amsfonts}%
\usepackage{amsthm}%
\usepackage{mathrsfs}%
\usepackage[title]{appendix}%
\usepackage{xcolor}%
\usepackage{textcomp}%
\usepackage{manyfoot}%
\usepackage{booktabs}%
\usepackage{algorithm}%
\usepackage{algorithmicx}%
\usepackage{algpseudocode}%
\usepackage{listings}%
\usepackage{indentfirst}
\usepackage[most]{tcolorbox}
\usepackage{tabularx}
\usepackage{makecell}
\usepackage{seqsplit}
\usepackage{hyperref}
\usepackage{filecontents}
\definecolor{darkgreen}{rgb}{0.0, 0.5, 0.0}
\usepackage{fontawesome}
\newcommand{\revise}[1]{\textcolor{black}{#1}}

\def \OURS{SciToolAgent}

\begin{document}

\title{\OURS{}: A Knowledge Graph-Driven Scientific Agent for Multi-Tool Integration}


\author[1,2]{\fnm{Keyan} \sur{Ding}}\email{dingkeyan@zju.edu.cn}
\equalcont{These authors contributed equally to this work.}
\author[2,4]{\fnm{Jing} \sur{Yu}}\email{yujing17@zju.edu.cn}
\equalcont{These authors contributed equally to this work.}
\author[2]{\fnm{Junjie} \sur{Huang}}\email{junjie6282@zju.edu.cn}
\author[2]{\fnm{Yuchen} \sur{Yang}}\email{yangyc\_2020@163.com}
\author*[3,2]{\fnm{Qiang} \sur{Zhang}}\email{qiang.zhang.cs@zju.edu.cn}
\author*[1,2,5]{\fnm{Huajun} \sur{Chen}}\email{huajunsir@zju.edu.cn}
\affil[1]{\orgdiv{College of Computer Science and Technology}, \orgname{Zhejiang University}, Hangzhou, China}
\affil[2]{\orgdiv{Zhejiang Key Laboratory of Intelligent Manufacturing for Functional Chemicals, ZJU-Hangzhou Global Scientific and Technological Innovation Center}, \orgname{Zhejiang University}, Hangzhou, China}
\affil[3]{\orgdiv{ZJU-UIUC Institute}, \orgname{Zhejiang University}, Haining, China}
\affil[4]{\orgdiv{The Polytechnic Institute}, \orgname{Zhejiang University}, Hangzhou, China}
\affil[5]{\orgname{State Key Laboratory of Ocean Sensing, Hangzhou, China}}

\abstract{ 
Scientific research increasingly relies on specialized computational tools, yet effectively utilizing these tools demands substantial domain expertise. While Large Language Models (LLMs) show promise in tool automation, they struggle to seamlessly integrate and orchestrate multiple tools for complex scientific workflows. Here, we present \OURS{}, an LLM-powered agent that automates hundreds of scientific tools across biology, chemistry, and materials science. At its core, \OURS{} leverages a scientific tool knowledge graph that enables intelligent tool selection and execution through graph-based retrieval-augmented generation. The agent also incorporates a comprehensive safety-checking module to ensure responsible and ethical tool usage. Extensive evaluations on a curated benchmark demonstrate that \OURS{} significantly outperforms existing approaches. Case studies in protein engineering, chemical reactivity prediction, chemical synthesis, and metal-organic framework screening further demonstrate \OURS{}'s capability to automate complex scientific workflows, making advanced research tools accessible to both experts and non-experts.
}

\maketitle

\section{Introduction}\label{sec1}

Scientific research increasingly depends on specialized computational tools that automate essential tasks ranging from data analysis to result visualization. While these tools have become indispensable for advancing scientific discovery, their growing complexity and diversity create substantial barriers to adoption. For example, Chemists routinely employ specific tools for molecular simulation, property prediction, and compound design. Beginner researchers may lack the technical expertise required to effectively utilize these powerful resources, potentially impeding scientific progress. To address this challenge, a promising approach involves leveraging artificial intelligence (AI) technologies to efficiently organize and orchestrate the use of scientific tools \cite{birhane2023science,schick2024toolformer,yang2024gpt4tools}.

Large language models (LLMs), as cutting-edge AI methods, have demonstrated unprecedented capabilities across diverse domains, from natural language understanding to complex reasoning tasks \cite{guo2023can,zhao2023survey,min2023recent}. 
Recent research has demonstrated promising advances in integrating LLMs with domain-specific scientific tools \cite{wang2024survey,ramos2024areview}. In chemistry, several pioneering systems—including Coscientist \cite{boiko2023autonomous}, ChemChat \cite{janakarajan2023language}, ChemCrow \cite{m2024augmenting}, and CACTUS \cite{mcnaughton2024cactus}—have enabled autonomous chemical research through LLM-tool integration. Similar progress has emerged in biological sciences, where systems like GeneGPT \cite{jin2024genegpt}, CRISPR-GPT \cite{huang2024crispr}, GenoAgent \cite{liu2024genotex}, and ProtAgents \cite{ghafarollahi2024protagents} have enhanced LLMs for specialized tasks such as RNA sequencing, gene editing, and protein discovery. The application of tool-augmented LLMs has further expanded to materials discovery (LLMatDesign \cite{jia2024llmatdesign}, ChatMOF \cite{kang2024chatmof}), electronic design (ChatEDA \cite{wu2024chateda}), and mechanics engineering (MechAgents \cite{ni2024mechagents}). These systems typically leverage in-context learning within established frameworks like ReAct \cite{yao2023react}, which combines LLM-based reasoning with tool execution. However, current approaches face two critical limitations: (1) they operate with a restricted set of tools (typically fewer than twenty), limiting their broader applicability, and (2) they often overlook crucial safety and ethical considerations in scientific research \cite{he2023control}.

Current agent frameworks that rely on naive in-context learning often struggle with complex scientific problems due to their inability to account for intrinsic dependencies among a wide array of tools \cite{liu2024toolnet,hao2024toolkengpt}. \revise{These dependencies are predominantly characterized by sequential relationships, where the output of one tool serves as the input for the next, necessitating a precise operational order.} The failure to account for these interdependent relationships frequently leads to suboptimal solutions and reduced efficiency when handling multi-step scientific workflows.
\begin{figure}[ht!]
    \centering
    \includegraphics[width=1\linewidth]{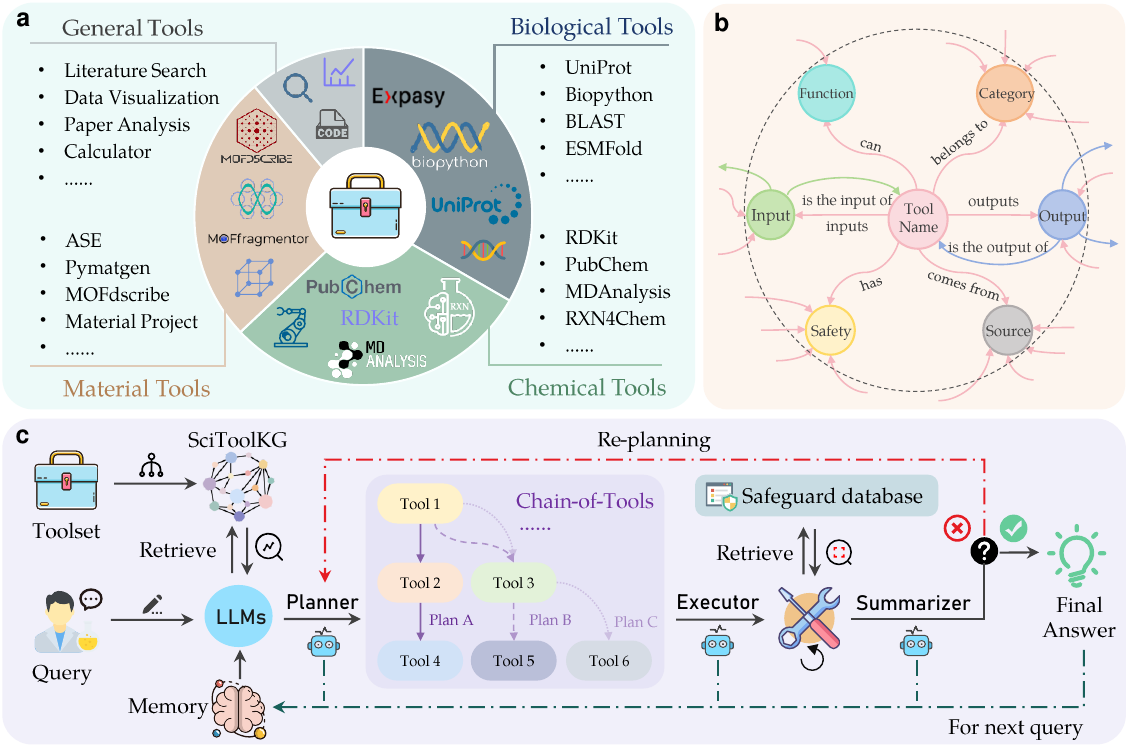}
    \caption{\textbf{Overview of \OURS{}.} \textbf{(a)} The toolset involved in our study, which includes general tools as well as frequently-used biology, chemistry, and material-related tools. \textbf{(b)} {The schema of SciToolKG, encompassing diverse information about each tool such as input/output formats, specific functionalities, safety levels, etc.} \textbf{(c)} The overall workflow of \OURS{}. Given a user query, an LLM-based Planner generates a chain of tools using a SciToolKG-based retrieve-augmented generation, followed by sequential implementation through an LLM-based Executor. A safety check module is then conducted to ensure ethical and safe solutions by retrieving a safeguard database. Finally, an LLM-based Summarizer concludes the implementation results, assesses the problem-solving process, and prompts the Planner to generate a new chain-of-tools if necessary. The final answers will be sent to the memory module as context for the next query.
}
    \label{fig:overview}
\end{figure}
In this study, we present \OURS{}, an agent framework that effectively integrates extensive and diverse scientific tools with LLMs (Fig. \ref{fig:overview}). Specifically, \OURS{} utilizes advanced LLMs as Planner, Executor, and Summarizer to autonomously plan, execute multiple tools, and summarize the solution for scientific tasks. 
\OURS{} introduces two key innovations: a comprehensive scientific tool knowledge graph (SciToolKG) that encodes relationships among hundreds of tools across biology, chemistry, and materials science, and an integrated safety module that ensures responsible scientific research. The SciToolKG enables informed tool selection and combination by explicitly modeling tool dependencies, prerequisites, and compatibility. Our safety module continuously monitors the execution pipeline to prevent potentially harmful outcomes, addressing a critical limitation in existing frameworks that often overlook the ethical implications of automated scientific discovery.

We evaluate \OURS{} using our scientific tool evaluation (SciToolEval) benchmark, comprising 531 diverse scientific problems that span multiple domains and complexity levels. Quantitative analysis demonstrates that \OURS{} achieves an overall accuracy of 94\%, surpassing state-of-the-art baselines by 10\%. We further validate \OURS{}'s effectiveness through case studies across four scenarios: protein design and analysis, chemical reactivity prediction, chemical synthesis and analysis, and metal-organic framework (MOF) material screening. These studies showcase \OURS{}'s capability to autonomously orchestrate complex multi-tool workflows while maintaining solution reliability and accuracy.

\section{Results}
\subsection{Overview of \OURS{}}

\OURS{} is an LLM-based agent designed to overcome the limitations of existing systems in scientific tool integration. While traditional frameworks often fail due to their naive in-context learning approaches that overlook tool interdependencies, \OURS{} addresses this challenge through a scientific tool knowledge graph (SciToolKG) that mediates between LLMs and scientific tools.
The toolset involved in \OURS{} includes general tools like search engines and code interpreters, as well as specialized scientific tools in biology, chemistry, and materials domains (Fig. \ref{fig:overview}a). 
To establish the intricate relationships among extensive tools, we manually construct a SciToolKG (Fig. \ref{fig:overview}b), encompassing diverse information about each tool. This SciToolKG plays a crucial role in the planning process, enabling LLMs to make well-informed decisions regarding tool selection and combination for optimal problem-solving.
The implementation workflow of \OURS{} is illustrated in Fig. \ref{fig:overview}c, which consists of three main components: a tool \textit{Planner}, a tool \textit{Executor}, and a solution \textit{Summarizer}, all of which are powered by LLMs. 
The Planner utilizes retrieve-augmented generation with SciToolKG to generate a chain of tools that can solve the given query. The Executor implements these tools sequentially and retries if errors occur. A retrieve-based safety check module is employed to identify the potential harmful responses within tools. The Summarizer compiles and synthesizes outputs from various tools to generate the final answer. In case of failure to solve the problem using the current plan, as judged by the summarizer, it will prompt the Planner to refine the chain of tools.
\begin{figure*}[ht!]
{
    \centering
    \includegraphics[width=1\linewidth]{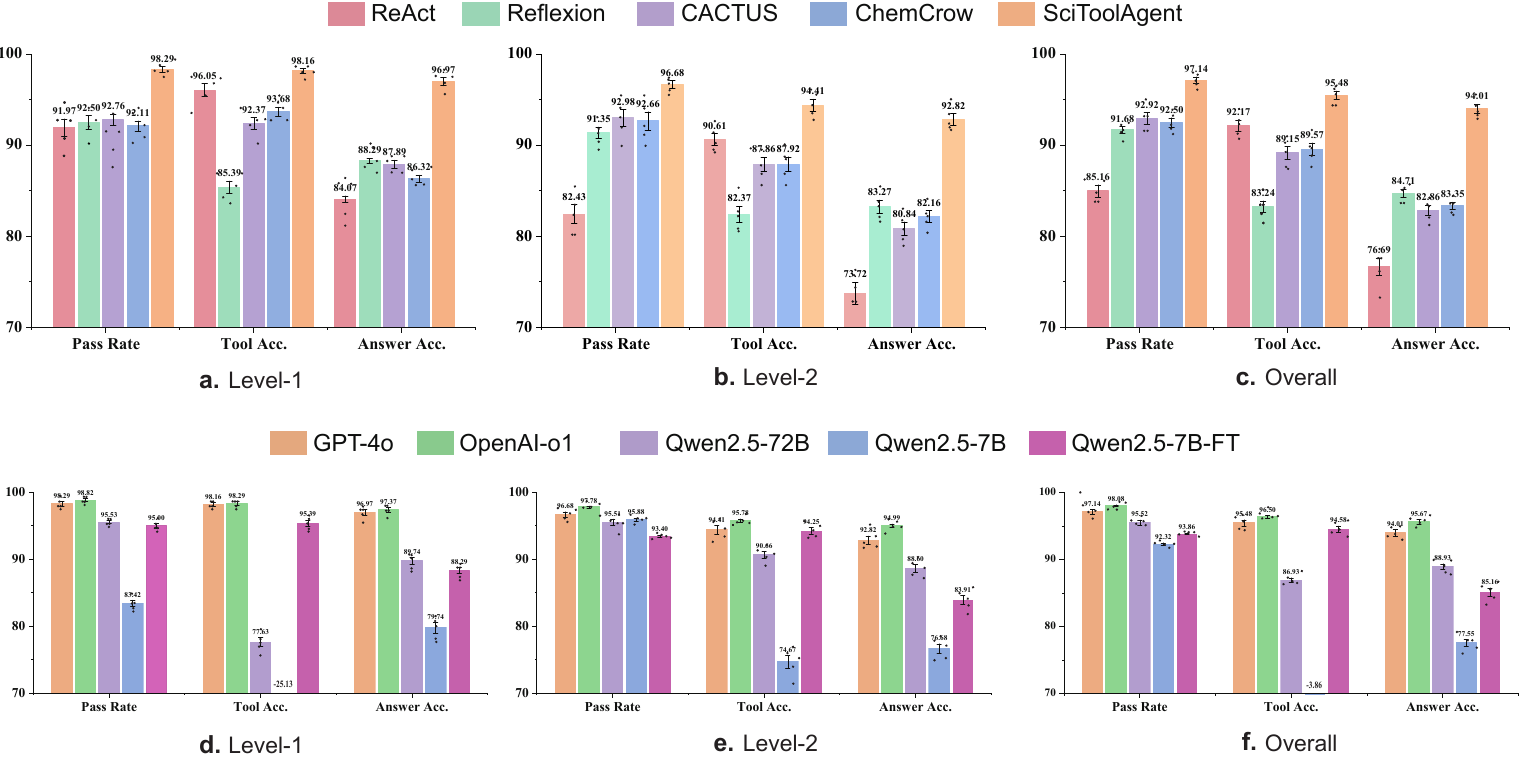}
    \vspace{-1em}
        \caption{\textbf{Comparison results of different agents and foundation models.} \textbf{(a)-(c)} Performance of different agents on SciToolEval, including Level-1 and Level-2, as well as overall performance. The proposed \OURS{} consistently outperforms the baseline approaches across all evaluation metrics and levels.  \textbf{(d)-(f)} Performance of different foundation models within \OURS{}. The OpenAI-o1 model achieves the best result, while the GPT-4o reaches an optimal trade-off between accuracy and cost. Data are presented as mean values. The number of experimental replicates is $n=5$, and the error bars are standard error of the mean.}
    \label{fig:result}
    }
\end{figure*}

\subsection{Performance on SciToolEval}

\textbf{Dataset:} To quantitatively evaluate the performance of \OURS{}, we curated SciToolEval, a comprehensive scientific tool evaluation dataset, comprising 531 diverse scientific questions across various fields, such as molecular property prediction, protein analysis, and material retrieval. The process of dataset construction is elaborated in Sec. \ref{sec:tooleval}.
The final dataset is divided into two levels: Level-1 includes 152 questions that can be addressed using a single tool, while Level-2 comprises 379 questions that require the use of multiple tools.

\textbf{Evaluation metrics:} To rigorously assess the effectiveness of \OURS{}, we utilize three distinct evaluation metrics: (1) Pass Rate, which quantifies the proportion of successfully completed queries; (2) Tool Planning Accuracy, which measures the alignment between the tool selection and sequencing generated by the agent and a reference plan verified by human experts; (3) Final Answer Accuracy, which compares the final solution generated by agents with the standard answer, assessing the correctness of the output. For both tool planning accuracy and final answer accuracy, we prompt GPT-4o to perform a similarity evaluation between the generated response and the ground truth.

\textbf{Baselines:} We compare \OURS{}'s performance against two state-of-the-art LLM-based tool agents including ReAct \cite{yao2023react}, Reflexion \cite{shinn2024reflexion}, CACTUS \cite{mcnaughton2024cactus}, and ChemCrow \cite{m2024augmenting}.
The ReAct framework involves an iterative process where LLMs generate a reasoning step (that is, thought) to determine how to tackle the problem and select a tool (that is, action) to take, followed by a result analysis (that is, observation) through LLMs. 
The Reflexion framework enhances ReAct by incorporating a feedback mechanism into the decision-making process, which includes self-evaluation, self-reflection and memory components. {ChemCrow and CACTUS are ReAct-based agents with domain-specific tools and templates for chemical synthesis and material science, respectively. While their default toolsets are limited, we adapt them with our full toolset for a fair comparison.
}

\textbf{Foundation models:} 
In our experiments, we utilize different foundation models for the Planner, Executor and Summarizer, including the proprietary model - OpenAI's GPT-4o (default) and o1, as well as the state-of-the-art open-source model - Qwen2.5-72B.
All of these models are accessed via official APIs with a standard configuration. 
Additionally, we explore Qwen2.5-7B, an open-source LLM with a smaller parameter scale but higher efficiency. Given its reduced inference capacity, we fine-tuned it using specific instructions for tool learning. This setup allows us to evaluate the cost-effectiveness of \OURS{} across models of varying capacities, ensuring that our framework can perform effectively while accommodating resource-constrained environments.

\textbf{Result Analysis:}
Fig. \ref{fig:result}a-c show the performance of five GPT-4o-based agents (ReAct, Reflexion, ChemCrow, CACTUS, and \OURS{}) across different levels on SciToolEval. \OURS{} outperformed the other agents in all evaluated metrics. For the task of using a single tool (Level-1), the agent needs to select a appropriate tool from the candidate tools, execute it correctly, and summarize the answer. For the task of using multiple tools (Level-2), the agent requires additional capabilities, including selecting multiple tools, generating a suitable plan, and executing them in a specific sequence. \OURS{} demonstrates a more significant advantage in solving multi-tool problems, outperforming ReAct and Reflexion by approximately 20\% and 10\%, respectively, in terms of final answer accuracy. Compared to ReAct-based agents ChemCrow and CACTUS, \OURS{} also achieves an absolute gain of 10–12\% on both levels. 
Empirically, we observe that ReAct and Reflexion exhibit limitations in tool planning and execution when confronted with complex multi-tool tasks and a large set of candidate tools. Both approaches lack a comprehensive, global task-planning strategy. Although Reflexion employs a feedback mechanism to improve pass rate and accuracy, its extensive trial-and-error process results in a lower tool planning accuracy. 
\OURS{} utilizes SciToolKG, significantly enhancing the accuracy of tool planning. Owing to the introduction of the chain-of-tools, \OURS{} can plan step-by-step tasks from a global perspective, reducing trial-and-error costs and improving its planning proficiency. During tool execution, the chain-of-tools helps the model focus on the current subtask, avoiding interference from irrelevant information, thus enhancing execution accuracy.

Fig. \ref{fig:result}d-f show the performance of five different foundation models within the \OURS{} framework. OpenAI o1 consistently outperforms other models across all metrics in both Level-1 and Level-2 tasks, followed by GPT-4o, which achieves slightly lower accuracy but maintains strong overall performance. Qwen2.5-72B performs competitively but demonstrates a slight drop in tool planning accuracy and final answer accuracy. In contrast, Qwen2.5-7B shows comparatively lower performance, especially in tool planning, indicating limitations in its capacity to manage complex tool integration. However, a finetuned version of Qwen2.5-7B (FT), trained using data generated from SciToolKG, exhibits significant improvement (+10\%) and achieves results closer to the higher-capacity models (Qwen2.5-72B). 
Notably, GPT-4o achieves an optimal trade-off between accuracy and cost, delivering near-top performance across all metrics while maintaining a significantly lower API cost compared to o1. This makes GPT-4o the default foundation model in our implementation of \OURS{}. 

\subsection{Case Studies}
To further validate the utility of \OURS{}, we conduct four real-world scientific research tasks with \OURS{}, including (1) protein design and analysis, (2) chemical reactivity prediction, (3) chemical synthesis and analysis, and (4) MOF materials screening.

\subsubsection{Protein design and analysis}\label{sec:case1} 
  
Protein design is crucial for advancing biological fields such as drug discovery, enzyme engineering, and synthetic biology. Designing proteins with desired properties is a complex task that typically requires the integration of multiple bioinformatics tools for sequence generation, folding prediction, and structural analysis. Traditional workflows necessitate substantial expertise to navigate and operate the diverse set of specialized tools effectively. \OURS{} streamlines this process by autonomously orchestrating the necessary tools to achieve comprehensive protein design and analysis. 

\begin{figure}[ht!]
    \centering
    \includegraphics[width=1\linewidth]{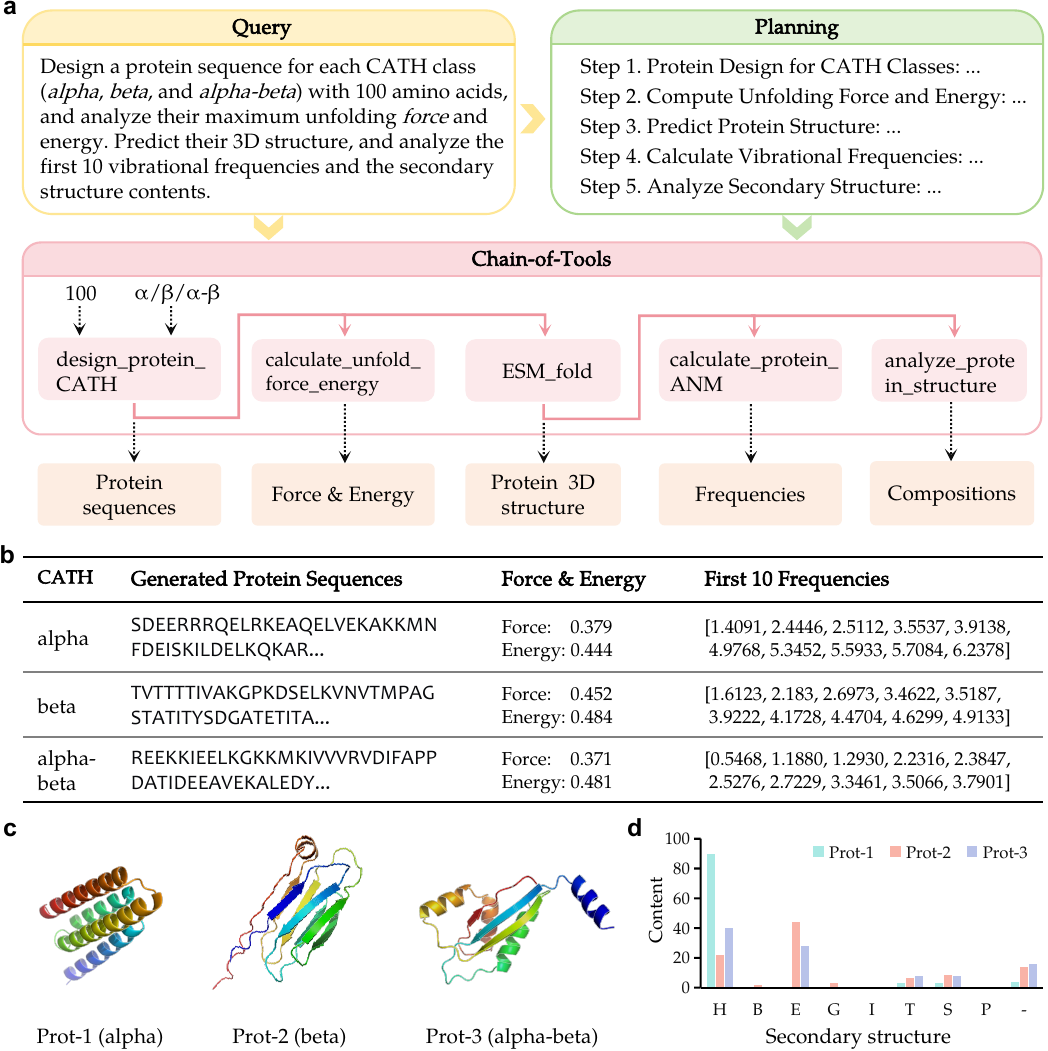}
    \vspace{-1em}
        \caption{\textbf{Workflow and results of protein design and analysis using \OURS{}.} \textbf{(a)} Overview of the input query, tool planning, chain-of-tools, and corresponding outcomes. \textbf{(b)} The generated protein sequences, the corresponding unfolding forces and energies, and the first 10 vibrational frequencies for each class. \textbf{(c)} The predicted 3D structures of the generated protein sequences. \textbf{(d)} The fractional content of secondary structure elements.
        }
    \label{fig:case1}
\end{figure}

In this case, we illustrate how \OURS{} manages a multi-step workflow for protein design.
Given a user query (Fig. \ref{fig:case1}a), the process begins with protein sequence generation, using Chroma \cite{ingraham2023illuminating} to create sequences based on predefined secondary structure types (for instance, $\alpha$-helix, $\beta$-sheet, or mixed). Next, unfolding force and energy calculations are performed using ProteinForceGPT \cite{ghafarollahi2024protagents} to assess the mechanical stability of the generated sequences. The protein structure is then predicted using ESMFold \cite{lin2023evolutionary}, followed by vibrational frequency computation with Anisotropic Network Model (ANM) \cite{atilgan2001anisotropy} implemented in ProDy \cite{bakan2011prody} to evaluate the dynamic stability of the structure. Finally, secondary structure analysis is carried out using the Dictionary of Secondary Structure of Proteins (DSSP) module from BioPython \cite{cock2009biopython} to confirm the structural elements.
The results (Fig. \ref{fig:case1}b-c) from this workflow demonstrate the capability of \OURS{} to autonomously orchestrate multiple tools and produce reliable outcomes. 
In contrast, the baseline methods ReAct and Reflexion failed to address this task effectively due to their incorrect or missing tools.

\subsubsection{Chemical reactivity prediction} \label{sec:case2}
\begin{figure}[ht!]
    \centering
    \includegraphics[width=1\linewidth]{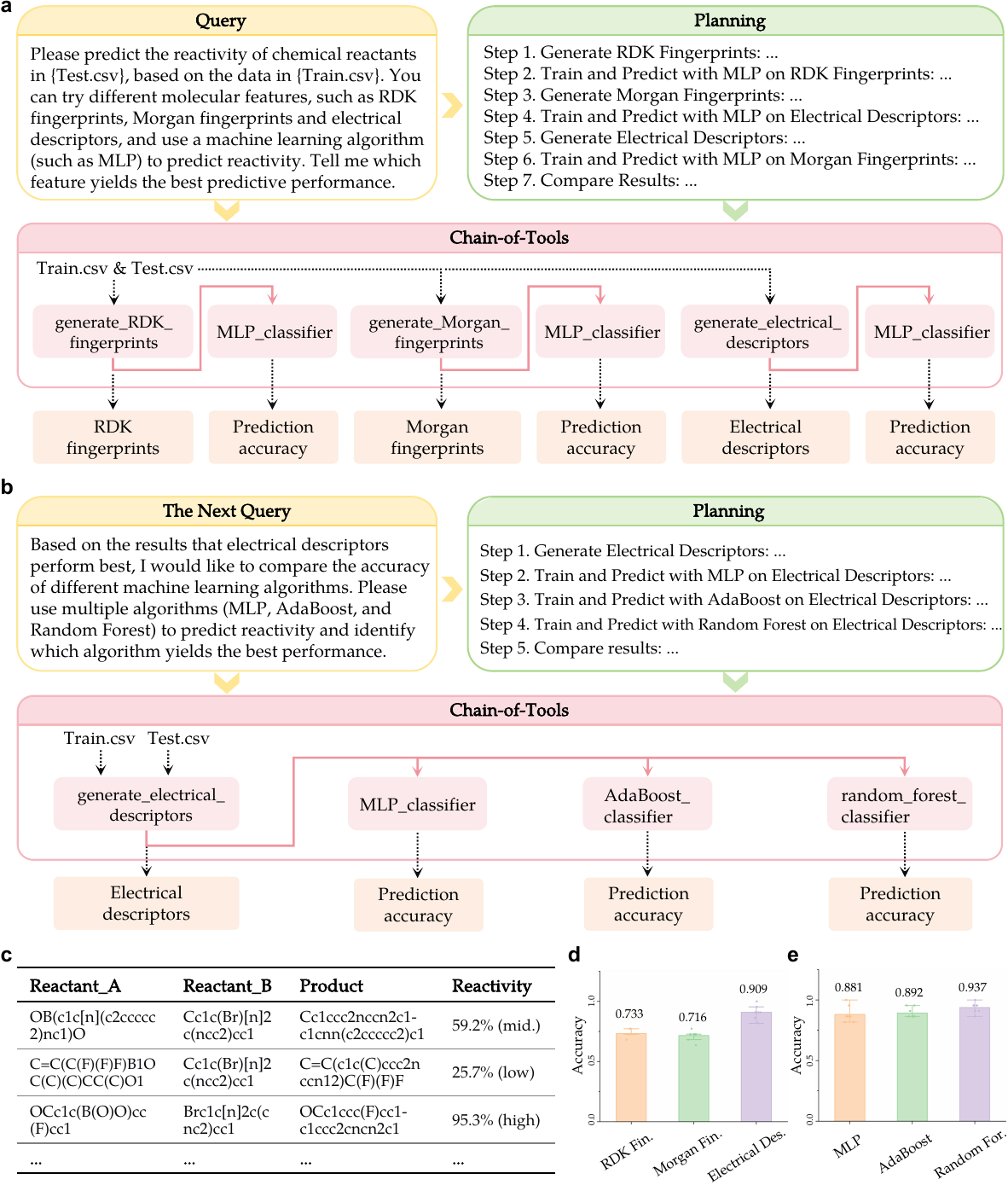}
    \vspace{-1em}
        \caption{\textbf{Workflow and results of machine learning-based chemical reactivity prediction using \OURS{}.} \textbf{(a)} and \textbf{(b)} Overview of the input query, tool planning, chain-of-tools, and corresponding outcomes. 
        \textbf{(c)} The dataset used for training and testing.
        \textbf{(d)} The prediction accuracy with different molecular features.
        \textbf{(e)} The prediction accuracy with different machine learning algorithms.
        Data are presented as mean values. The number of experimental replicates is n=8, and the error bars are standard error of the mean.} 

    \label{fig:case2}
\end{figure}

Predicting chemical reactivity is a critical aspect of drug design and organic synthesis. Accurate predictions of how chemical compounds will react with one another can accelerate the development of new compounds and reduce the need for labor-intensive experimental processes. 
In this case, we employ \OURS{} to predict the reactivity of amide condensation reactions, a crucial class of reactions in organic chemistry. Specifically, we aim to predict the reactivity of various chemical substrates based on molecular features and identify the most effective machine learning algorithms for this task. The dataset (Fig. \ref{fig:case2}c) used for this study comprises different combinations of reactants along with the corresponding reaction products, and the conversion rate of the product, quantified by the liquid chromatography area percent (LCAP) at UV 254 nm. The goal is to classify the reactivity of these substrate combinations into three categories: low, medium, and high activity.
The workflow (Fig. \ref{fig:case2}a) begins with an initial query prompting \OURS{} to predict reactivity using various molecular features, such as RDK fingerprints, Morgan fingerprints, and electrical descriptors, and to test a machine learning algorithm, like the Multi-Layer Perceptron (MLP), to determine which feature provides the best predictive performance. \OURS{} plans the steps accordingly, first generating each type of fingerprint and descriptor, then training and testing the MLP classifier on each feature set. The results (Fig. \ref{fig:case2}d) indicate that electrical descriptors outperform the other features in predicting reactivity.
Based on this finding, a follow-up query asks \OURS{} to further explore the performance of different machine learning algorithms, specifically comparing MLP, AdaBoost, and Random Forest classifiers, using only electrical descriptors as input features (Fig. \ref{fig:case2}b). \OURS{} then generates a new plan to test each algorithm sequentially on this feature set, evaluating their predictive accuracy in classifying reactivity. By comparing the outcomes, \OURS{} determines which algorithm is most effective with the electrical descriptors. The results (Fig. \ref{fig:case2}e) reveal that the Random Forest algorithm yields the highest prediction accuracy among the tested algorithms, making it the optimal choice for this task. 
Notably, ReAct encountered tool redundancy and Reflexion encountered hallucinations in completing this task.

\subsubsection{Chemical synthesis and analysis} \label{sec:case3}
Chemical synthesis is a cornerstone of chemistry, pivotal for the design and creation of novel compounds, particularly in pharmaceutical research. Accurately predicting reaction outcomes and understanding the properties of synthesized molecules are crucial steps for efficient and safe drug design.  In this case, we demonstrate how \OURS{} automates a synthesis and analysis pipeline, encompassing reaction prediction, product characterization, intellectual property assessment, and safety evaluation.

\begin{figure}[!th]
    \centering
    \includegraphics[width=1\linewidth]{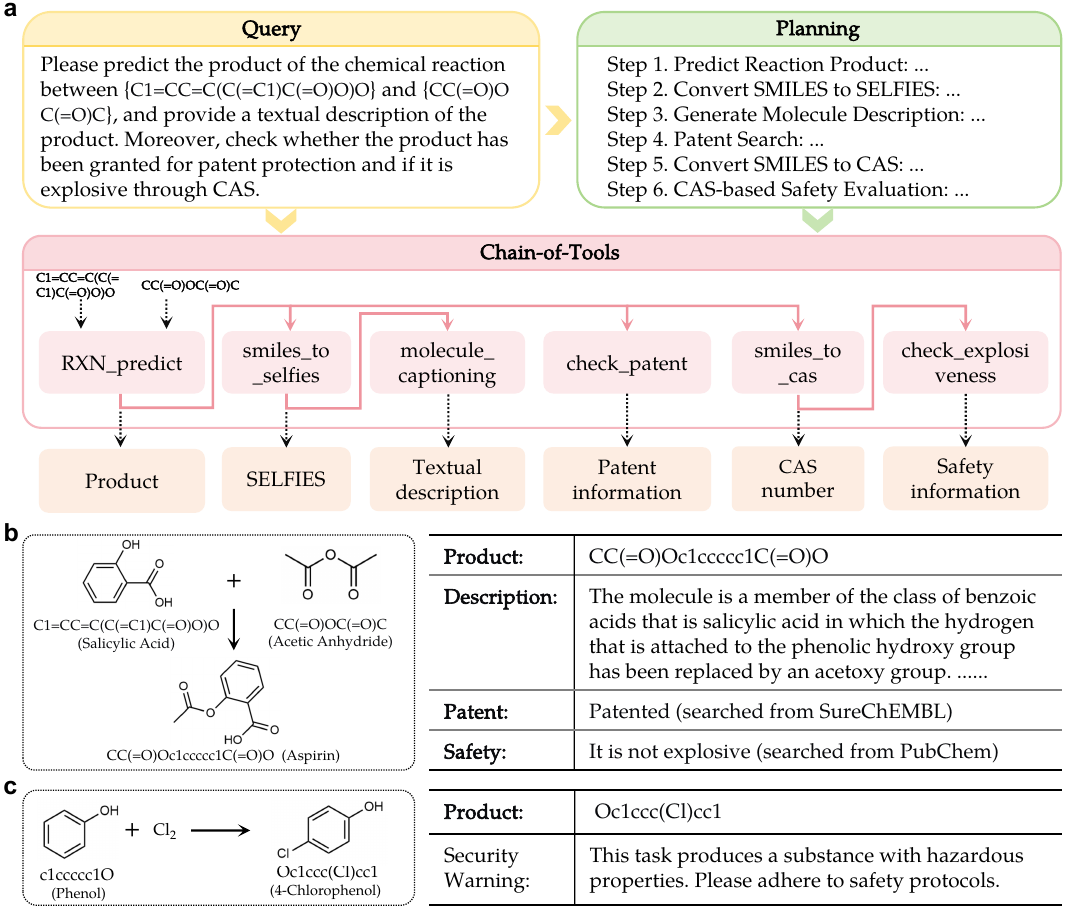}
    \vspace{-1em}
        \caption{\textbf{Workflow and results of chemical synthesis and analysis using \OURS{}.}
        \textbf{(a)} Overview of the input query, tool planning, chain-of-tools, and corresponding outcomes. 
        \textbf{(b)} Results of the chemical reaction between Salicylic Acid and Acetic Anhydride, including the product and the corresponding description, patent, and safety information. 
        \textbf{(c)} Results of the chemical reaction between Phenol and Chlorine, issuing a security warning due to the product's toxicity.
        }
    \label{fig:case3}
\end{figure}

The first task involves predicting and analyzing the outcome of an esterification reaction between salicylic acid and acetic anhydride. 
\OURS{} autonomously plans and executes this multi-step process by orchestrating the necessary tools in sequence, as shown in Fig. \ref{fig:case3}a. Specifically, \OURS{} first employs the RXN-for-chemistry \cite{schwaller2019molecular} tool to predict the reaction product (denoted by SMILES). Following this, the agent converts the SMILES notation of the product into the SELFIES format for compatibility and uses a molecule captioning tool (BioT5+ \cite{pei2024biot5+}) to generate a textual description. Next, the agent conducts a patent search using a patent check tool across the SureChEMBL database \cite{papadatos2016surechembl}. Finally, for explosiveness assessment, the agent converts the SMILES notation to a CAS number and retrieves relevant safety information from PubChem \cite{kim2016pubchem}. The obtained results are summarized in Fig. \ref{fig:case3}b.

Additionally, we conducted a similar query for another reaction: the chlorination of phenol, where phenol reacts with chlorine (Fig. \ref{fig:case3}c). \OURS{} follows a comparable workflow for this query, beginning with reaction prediction. However, the internal safety check module in \OURS{} identifies the product (4-chlorophenol) as a hazardous compound, and issues a security warning indicating that the resulting product is toxic and requires careful handling and specific safety precautions. This built-in safety check is crucial, as it alerts researchers to potential risks early in the workflow, helping to prevent accidental exposure to toxic substances and ensuring that safety protocols are adhered to throughout the experimental process. Conversely, neither ReAct nor Reflexion incorporated safety checks during the synthesis process, which led to substantial risks in chemical experimentation. While both methods were able to predict reaction outcomes, they failed to identify the toxic nature of the product in the chlorination of phenol case.

\subsubsection{MOF materials screening} \label{sec:case4}

Metal-Organic Frameworks (MOFs) are a class of crystalline porous materials with versatile applications in areas such as gas storage, catalysis, drug delivery, and separation processes. The identification of optimal MOFs for specific scenarios often involves screening materials based on multiple performance criteria, such as high thermal stability and efficient adsorption capacity. In this case, we demonstrate how \OURS{} can streamline the MOF screening process by automating the selection and analysis of MOFs based on predefined criteria.
\begin{figure}[ht!]
    \centering
    \includegraphics[width=1\linewidth]{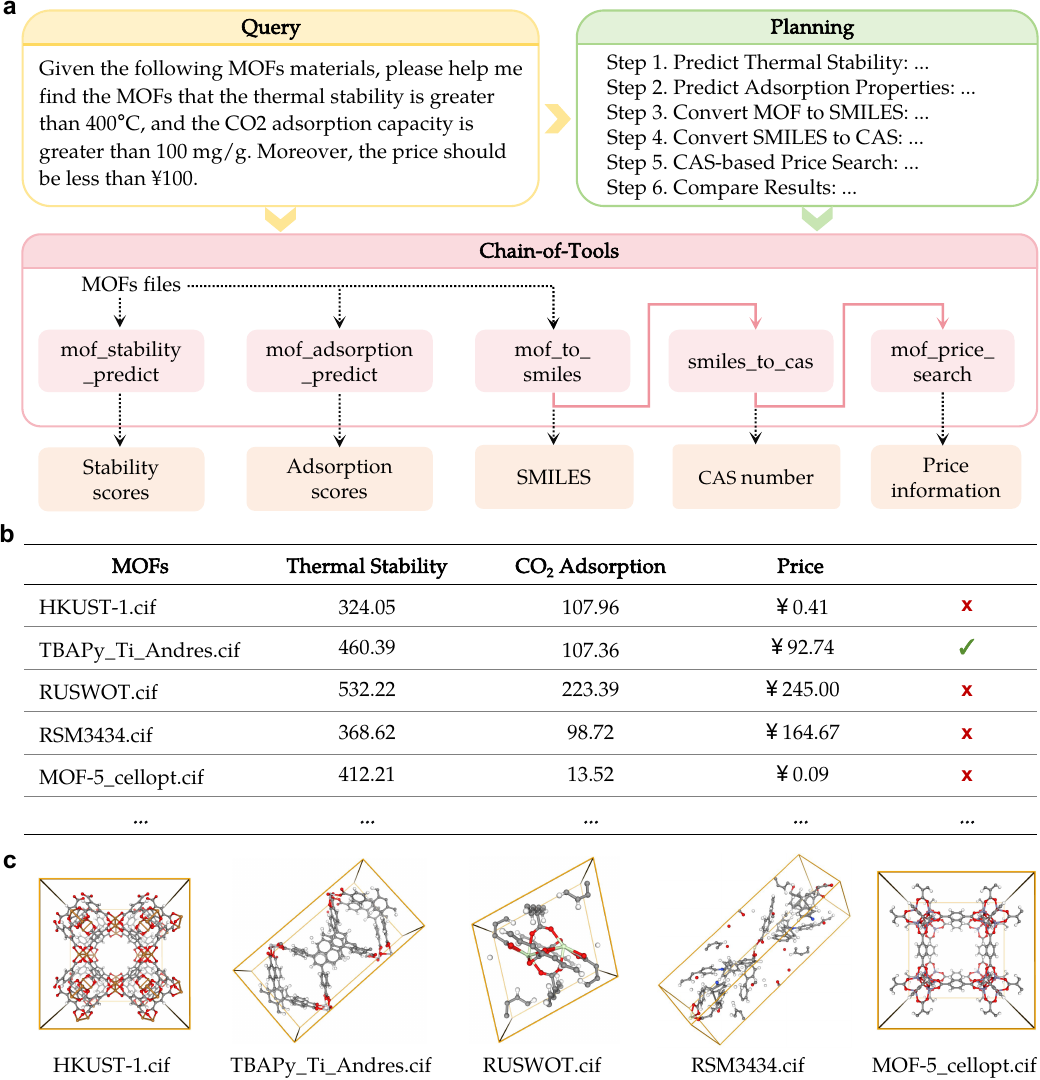}
    \vspace{-1em}
        \caption{\textbf{Workflow and results of MOF materials screening using \OURS{}.} \textbf{(a)} Overview of the input query, tool planning, chain-of-tools, and corresponding outcomes. \textbf{(b)} Summary of the obtained results showing the thermal stability, CO$_2$ adsorption capacity, and market price for each MOF. The green checkmarks indicate MOFs that meet all specified criteria. \textbf{(c)} Visualization of these MOF structures. 
        }
    \label{fig:case4}
\end{figure}

Given a batch of MOFs from MOFXDB \cite{bobbitt2023mofx}, the workflow (Fig. \ref{fig:case4}a) begins with a user query that prompts \OURS{} to find MOFs with thermal stability above 400$^\circ$C, CO$_2$ adsorption capacity over 100 mg/g, and a price under \textyen 100. In response, \OURS{} generates a plan and executes a chain of tools to filter MOFs based on these criteria. Specifically, the agent first assesses the thermal stability of each MOF by using a machine learning-based predictive model (MOFSimplify \cite{nandy2022mofsimplify}) that calculates thermal stability scores. Next, the agent predicts their CO$_2$ adsorption capacities using a molecular simulation software (RASPA2 \cite{dubbeldam2016raspa}). Last, the agent inquires about their pricing information. As the pricing retrieval tool requires CAS numbers as input, it first converts the structural data of MOFs (provided as CIF files) into SMILES format and subsequently translates these SMILES strings into CAS numbers. With the obtained CAS numbers, \OURS{} accesses the commercial chemical database to retrieve market price information.

Fig. \ref{fig:case4}b summarizes the partial results obtained from this automated screening workflow. For instance, by filtering MOFs based on thermal stability, CO$_2$ adsorption capacity, and price, \OURS{} successfully identified \texttt{TBAPy\_Ti\_Andres.cif} as a highly promising candidate that met all predefined criteria. 
The visualization of selected MOF structures (Fig. \ref{fig:case4}c) further supports the analysis by highlighting structural elements contributing to stability and adsorption properties, offering valuable insights for subsequent experimental validation.
In contrast, ReAct encountered input errors and Reflexion encountered hallucinations when attempting to address this task.

\section{Discussion}
\OURS{} is an LLM-powered agent designed to address the challenges of integrating extensive scientific tools for advanced research across multiple domains. 
The primary advantage of \OURS{} lies in its integration of diverse scientific tools through SciToolKG, which captures the intricacies of tool dependencies, input/output formats, and application contexts. Previous LLM-based frameworks have been restricted by their limited toolsets and simple task-planning strategies. By contrast, \OURS{} can dynamically create a chain of tools tailored to the specific requirements of each scientific task. 
This capability allows researchers to delegate repetitive or computationally intensive steps to \OURS{}, making scientific inquiry more accessible and efficient for both domain experts and non-experts.

One potential limitation of \OURS{} lies in the manual construction of the SciToolKG knowledge graph. Although SciToolKG captures relationships and dependencies effectively, its scalability is constrained by the effort required to curate and update tool information. Automated approaches, such as using knowledge extraction techniques from scientific literature or metadata from tool documentation, could further enhance the scalability and granularity of SciToolKG. Additionally, expanding the knowledge graph to incorporate emerging tools and resources will be essential as scientific research continues to evolve. To facilitate extensibility, we provide standardized APIs and templates to support third-party tool integration with minimal effort. Future versions will also include GUI-based registration to reduce the barrier for domain experts without programming experience.

Another challenge lies in the reliance on the underlying LLM's capabilities to perform effectively across varied scientific tasks. While proprietary models such as GPT-4o demonstrate strong performance, they may be financially and technically inaccessible for all researchers, especially in resource-limited settings. Our experiments with open-source models indicate that fine-tuning with domain-specific data can enhance performance and partially bridge the gap between open-source and proprietary alternatives. However, even with fine-tuning, Qwen2.5-7B-FT still lags behind GPT-4o, especially in complex tool planning and multi-step reasoning.

While challenges related to the scalability of SciToolKG and the reliance on proprietary LLM models remain, \OURS{} offers a robust foundation for automating complex scientific workflows. 
Future work will focus on automating the knowledge graph’s maintenance, integrating more tools, and enhancing open-source LLM capabilities to further democratize access to advanced scientific research. Ultimately, \OURS{} showcases the potential of LLM-driven agents to streamline and empower scientific discovery, making sophisticated tools accessible to a broader audience.

\section{Methods}
This section presents the methodology of \OURS{}, including the collection of scientific tools, the construction of SciToolKG and SciToolEval based on these tools, and the implementation details of Planner, Executor, Summarizer, as well as the LLMs behind them.
 
\subsection{Collection of Scientific Tools}\label{sec:tools}
The collection of scientific tools for \OURS{} follows a systematic process aimed at assembling a comprehensive, domain-specific, and functionally diverse toolset.  Initially, we identified key scientific domains that would benefit most from LLM integration, including biology, chemistry, and materials science. We then curated a list of tools used frequently in these domains.
The current toolset in \OURS{} includes over 500 tools, spanning a wide range of functionalities. For instance, in biology, we included sequence alignment tools like BLAST \cite{BLAST}, protein structure prediction models such as ESMFold \cite{lin2023evolutionary}, and gene editing software like CRISPR-related tools. In chemistry, our toolset features molecular dynamics simulations, cheminformatics libraries like RDKit \cite{rdkit}, and compound databases such as PubChem \cite{kim2016pubchem}. For materials science, we integrated tools for crystal structure prediction, materials property databases, and MOF property predictors such as MOFSimplify \cite{nandy2022mofsimplify}. 

\subsection{Construction of SciToolKG}\label{sec:toolkg}
The Scientific Tool Knowledge Graph (SciToolKG) serves as the backbone of \OURS{}, providing a structured representation of the relationships, dependencies, and operational details of a vast array of scientific tools. 
Formally, SciToolKG is represented as a directed graph \( G = (V, E) \), where \( V \) represents a set of entities (nodes) and \( E \) represents a set of relations (edges) between these entities.
Specifically, the tool nodes represent individual scientific tools, and the attribute nodes represent various tool properties and metadata.
The edges captures the semantic relationships between the tool node and the attribute node.

The construction of SciToolKG followed a three-step process: tool characterization, schema development, and graph population. 
The initial phase involved defining a set of attributes for each tool, including its purpose, specific functionalities, input/output formats, categories, sources, and safety levels. These attributes were mainly derived from tool documentation. 
Then, a hierarchical schema was developed to model these attributes logically, including nodes representing tools and their properties and edges denoting relationships such as functional dependencies.
The final stage involved populating the KG by encoding tool-specific metadata into a structured graph with triplets. In particular, relationships between tools—such as shared input/output formats or sequential task dependencies—were explicitly captured, creating a rich, interconnected graph. To ensure accuracy and usability, the KG underwent iterative refinement through expert validation.

\subsection{Construction of SciToolEval}\label{sec:tooleval}

To evaluate \OURS{}'s performance, we constructed a comprehensive Scientific Tool Evaluation (SciToolEval) dataset that encompasses a diverse array of real-world scientific tasks. This dataset serves as the first testing ground that can quantitatively assess agents' capabilities across various scientific domains.  The process of constructing SciToolEval involves the following steps.

\textbf{(1) Tool selection and question generation:} The first step is selecting the appropriate tools and generating meaningful questions. For single tool calling scenarios, we include a detailed description of the tool in the prompt and instruct GPT-4o to generate relevant questions. This ensures that the generated questions are directly actionable and pertinent to the tool's functionality.

\textbf{(2) Tool execution and answer generation:} Once the questions are generated, we proceed to tool execution and answer generation by utilizing the ReACT framework in conjunction with GPT-4o. ReACT \cite{yao2023react} has demonstrated its effectiveness in handling a limited number of tools. Therefore, only the tools relevant to the question are fed to ReACT, improving the accuracy of tool invocation.

\textbf{(3) Manual review:} The final step involves manual review to ensure the quality of generated answers. We invite three domain-specific experts to meticulously review the questions, tool usage, and answers to verify their correctness and significance.

\revise{Finally, SciToolEval consists of 531 questions associated with the required tools and their corresponding answers.} These questions cover single and multiple tools for problem-solving, and they are categorized into different levels based on difficulty: \textbf{Level-1} involves simple information queries using a single tool; \textbf{Level-2} includes sequential and parallel use of multiple tools, requiring basic planning, reasoning, and summarization skills.

\subsection{Implementation of \OURS{}}\label{sec:implementation}
\OURS{}'s architecture revolves around three core components powered by LLMs: Planner, Executor, and Summarizer (Fig. \ref{fig:overview}c). Each component plays a distinct role in the problem-solving workflow by leveraging the strengths of LLMs.

\subsubsection{Planner}\label{sec:planner}

The Planner is responsible for devising a strategy to solve the given query. It leverages LLMs to interpret the question, query the SciToolKG, and generate a plan of chain-of-tools. Specifically, a retrieve-augmented generation approach based on the SciToolKG is proposed to identify and sequence the appropriate tools needed. 
The process begins with the LLM querying the SciToolKG based on the input question, followed by the steps described below.

\textbf{(1) Full-graph retrieval:} We first retrieve the most relevant $k$ tools for the given question in the full graph of SciToolKG. This is achieved by calculating the semantic similarity between the question and all of the triples involved in tool functions within the SciToolKG. The similarity score helps in identifying tools that are most likely to be useful for the given query. In mathematical, let $T_i$ represent a tool in the SciToolKG $\mathcal{G}$ and $S(q, T_i)$ denote the semantic similarity score between the question $q$ and the tool $T_i$. This retrieval process can be formulated as:
\begin{align}
    \mathbb{T}_{\text{full}} = \text{top-}k \{T_i \mid S(q, T_i), \forall T_i \in \mathcal{G} \}.
\end{align}

\textbf{(2) Sub-graph exploration:} Once the initial set of tools is retrieved, we explore all sub-graphs (that is, $d$-hop neighborhoods) of these tools in SciToolKG. 
The purpose of this step is to identify any additional tools that might be required in conjunction with the initially retrieved tools to solve the query.
Mathematically, for each retrieved tool $T_i$, {we calculate the semantic similarity between $q$ and the tools $T_i \oplus T_j$, where $\oplus$ denotes semantic concatenation that combines the textual metadata of \( T_i \) and \( T_j \). }
Here, $T_j$ is the tool in the $d$-hop neighborhoods of $T_i$. The top-$k$ tools with the highest similarity are obtained by:
\begin{align}
   \mathbb{T}_{\text{sub}} = \text{top-}k \{T_j \mid S(q, T_i \oplus T_j), \forall T_j \in \mathcal{G}^{T_i}_{\text{sub}} \}.
\end{align}

\textbf{(3) Tool combination and ranking:} By employing full-graph retrieval and sub-graph exploration for each query, we obtain a total of $k^2$ tools. To optimize the utilization of these tools, we sort them according to the combined similarity score $S' = S(q, T_i) \times  S(q, T_i \oplus T_j)$, prioritizing tools that are not only relevant to the question but also complementary to each other. We then select the top-$n$ tools ($n\leq k^2$) based on the combined similarity:
\begin{align}
\mathbb{T}_{\text{comb}} = \text{top-}n \{T_{i} \mid S(q, T_i) \times  S(q, T_i \oplus T_j), \forall T_j \in \mathcal{G}^{T_i}_{\text{sub}}, \forall T_i \in \mathcal{G} \}.
\end{align}

\textbf{(4) Chain-of-tools generation:} {Based on the selected tools and the neighborhood information from SciToolKG}, we prompt LLMs to generate a chain-of-tools that outlines the tools required to solve the query. This chain is optimized to ensure that the tools are used in the most effective order, leveraging their dependencies and functionalities. The final chain-of-tools can be represented as:
\begin{align}
   \mathbb{T}_{\text{chain}} = \{T_1 \rightarrow T_2 \rightarrow  \ldots \rightarrow  T_m\} \in \mathbb{T}_{\text{comb}}.
\end{align}

Unless otherwise specified, the parameters for retrieval are set as follows: $d=3$, $k=5$, and $n=10$. For measuring semantic similarity $S$, we calculate the cosine distance between two text embedding vectors obtained from a pretrained embedding model. 

\subsubsection{Executor}\label{sec:executor}
The Executor aims to ensure the effective implementation of the planned chain-of-tools. This component is designed to handle tool inputs, manage execution processes, and address any errors that may arise during execution. It also integrates a robust safety module to monitor and control potentially hazardous inputs or outputs. 
The execution process involves the four steps as follows.
 
\textbf{(1) Input preparation:}
The Executor uses LLMs to extract the required input parameters for the current tool in the chain. This involves parsing the question, understanding the context, and identifying the necessary data to feed into the tool. Inputs are formatted according to the specifications of the tool recorded in SciToolKG. This may include converting data types, structuring the input in a specific format, or integrating multiple inputs.

\textbf{(2) Tool execution:}
The Executor invokes the tool with the prepared inputs, which involves interfacing with the tool’s API. The execution process is actively monitored to track progress, capture outputs, and detect any anomalies or errors. The Executor logs execution details to ensure traceability. Once the tool has been executed, the outputs are captured, and further processed as needed to be compatible with the subsequent tools in the chain. 

\textbf{(3) Error handling and retries:}
The Executor is equipped with protocols to detect errors during tool execution. Errors can occur due to incorrect inputs, tool malfunctions, or other unforeseen issues. When an error is detected, the Executor adjusts the inputs based on predefined rules or heuristics to rectify the issue. 

\textbf{(4) Retrieve-based safety check module:}
In parallel with the execution process, the Executor incorporates a retrieve-based safety check module. This module evaluates each step's inputs and outputs against a set of predefined safety criteria to prevent the generation of harmful substances or unethical use of scientific tools.
To this end, we first construct a comprehensive safeguard database that includes hazardous compounds collected from PubChem and toxic proteins collected from UniProtKB. 
By comparing the outcomes with the substances contained in this database, we can assess the potential harmfulness of molecules or protein sequences generated during execution.

Specifically, given a molecule \(x\), the safety check module calculates the feature similarity between \(x\) and all entries in the safeguard database \(D\). The similarity is quantified using the average of Tanimoto coefficient, Dice coefficient and Cosine coefficient, which measure the degree of similarity between molecular fingerprints \cite{bajusz2015tanimoto}:
\begin{align}
    \hat{S}_{\text{mol}}(x, D) = \max_{\forall y_i \in D} \frac{1}{3}\left( \text{Tanimoto}(x, y_i) + \text{Dice}(x, y_i) + \text{Cosine}(x, y_i) \right).
\end{align}
For proteins, we use the Smith-Waterman algorithm \cite{smith1981identification} for pairwise local alignment between the given sequence and entries in the safeguard database, and the percentage of matching serves as the similarity score:
\begin{align}
    \hat{S}_{\text{prot}}(x, D) = \max_{\forall y_i \in D} \text{Smith-Waterman}(x, y_i). 
\end{align}
If $\hat{S}(x, D)$ exceeds a certain threshold $\delta=0.95$, indicating a close to known hazardous or toxic entities, the safety module flags this output as potentially dangerous. 
To ensure the efficiency of execution, only the tools marked with high risks in SciToolKG are required to pass through the safety check module.

\subsubsection{Summarizer} \label{sec:summarizer}
After execution, the Summarizer compiles and synthesizes outputs from various tools to generate the final answer, ensuring coherence and accuracy. It also assesses the success of the problem-solving process and, if necessary, prompts the Planner to refine the chain-of-tools for optimal results. 
The implementation of the Summarizer involves several key steps.

\textbf{(1) Synthesis of outputs:}
The Summarizer collects the outputs from all the tools executed in the chain and integrates them into a cohesive response. This involves 1) combining the outputs from different tools and ensuring that all relevant information is included, 2) verifying that the outputs from different tools align with each other and do not contain conflicting information, and 3) structuring the final answer in a logical and easy-to-understand format.

\textbf{(2) Iterative refinement:} 
If the initial chain-of-tools fails to produce satisfactory results, as determined by the Summarizer, it prompts the Planner to refine the plan. This refinement involves identifying reasons for any failures or suboptimal outcomes, suggesting modifications to the chain-of-tools (such as adding, removing, reordering, or retrieving tools), and repeating the execution process with the refined chain-of-tools to achieve improved results.

\subsubsection{Foundation models} \label{sec:llm}
The foundation model behind the Planner, Executor and Summarizer in \OURS{} are large language models (LLMs).
Our experiments have tried different LLMs, including the well-known proprietary model - OpenAI's GPT-4o, as well as the state-of-the-art open-source models - Qwen2.5-72B. 
Moreover, we fine-tuned Qwen2.5-7B, an open-source LLM with a smaller parameter scale but higher efficiency, using specific instructions for tool learning and deployed it locally on a GPU server. 

\textbf{(1) Instruction generation:}
To enhance the performance of Qwen2.5-7B, we generate specific instructions tailored to scientific tool learning. These instructions guide the model in understanding the functionalities of various tools and their appropriate usage. 
{The instruction generation process is similar to the construction of the SciToolEval dataset, utilizing SciToolKG and GPT-4o for automated generation (encompassing tool selection, question generation, tool execution, and answer generation). This approach yields a large number of instructions, although they are not manually reviewed.}

\textbf{(2) Fine-tuning LLMs with instructions:}
We fine-tuned Qwen2.5-7B using LoRA (Low-Rank Adaptation \cite{hu2021lora}), leveraging the constructed instruction dataset to enhance the model's capacity for tool-specific planning and execution. Separate fine-tuning was conducted for the Planner, Executor, and Summarizer, with task-specific instructions ensuring the optimization of each component's functionality.
The training loss is measured using the next token prediction loss \cite{vaswani2017attention}, defined as:
\begin{align}
   L = -\sum_{t=1}^{T} \log P(y_t \mid y_{<t}, X; \theta),
\end{align}
where \( y_t \) represents the target token at time step \( t \), \( X \) is the input text, and \( \theta \) denotes the model parameters.
This fine-tuning process significantly enhances the instruction following capability of Qwen2.5-7B, making it a cost-effective choice for implementing the Planner, Executor and Summarizer. 

\clearpage

\section*{Data Availability}
The toxic compound data were obtained from PubChem (\url{https://pubchem.ncbi.nlm.nih.gov/#query=toxic&tab=compound}), and the toxic protein data were sourced from UniProtKB (\url{https://www.uniprot.org/uniprotkb?query=toxic&facets=reviewed%3Atrue}).

\section*{Code Availability}
The source code of \OURS{} can be found at GitHub (\url{https://github.com/hicai-zju/scitoolagent}).


\end{document}